\documentclass{article}

\usepackage[final]{neurips_2020}

\usepackage[utf8]{inputenc} 
\usepackage[T1]{fontenc}    
\usepackage{hyperref}       
\usepackage{url}            
\usepackage{booktabs}       
\usepackage{amsfonts}       
\usepackage{nicefrac}       
\usepackage{microtype}      

\usepackage{graphicx}
\usepackage{algorithm}
\usepackage[noend]{algpseudocode}
\usepackage{xcolor}
\usepackage{graphicx}

\title{Evaluate On-the-job Learning Dialogue Systems and a Case Study for Natural Language Understanding}

\author{%
  Mathilde Veron$^{1,2}$, Sophie Rosset$^{1}$, Olivier Galibert$^{2}$, Guillaume Bernard$^{2}$\\
  $^{1}$LIMSI, Université Paris-Saclay, CNRS - \texttt{firstname.name@limsi.fr} \\
  $^{2}$LNE - \texttt{firstname.name@lne.fr}
}

\begin{document}

\maketitle

\begin{abstract}
    On-the-job learning consists in continuously learning while being used in production, in an open environment,
    meaning that the system has to deal on its own with situations and elements never seen before.
    The kind of systems that seem to be especially adapted to on-the-job learning are dialogue systems, since they can
    take advantage of their interactions with users to collect feedback to adapt and improve their components over time.
    Some dialogue systems performing on-the-job learning have been built and evaluated but no general methodology has yet
    been defined.
    Thus in this paper, we propose a first general methodology for evaluating on-the-job learning dialogue systems.
    We also describe a task-oriented dialogue system which improves on-the-job its natural language component through
    its user interactions.
    We finally evaluate our system with the described methodology.
\end{abstract}

\section{Introduction}
\label{sec:intro}

When deploying a system and making it available to real users, any developer wishes that the system would continue learning on its own when being used.
Such a system would be able to continuously improve its components, adapt to unforeseen elements and situations and extend its domain without intervention from developers or domain experts.
Especially, developers would avoid manual log analysis and system modifications.
This capability is related to Lifelong Learning (LL) \citep{10.1007/978-3-642-79629-6_7, Chen:2016:LML:3086758} and
never-ending learning \citep{Carlson2010TowardAA, 10.1145/3191513}.
Recently \citet{bing_on_the_job_2020} explicitly incorporated on-the-job learning in the definition of LL.
It consists in continuously learning while being used in production, in an open environment, meaning
that the system has to deal on its own with situations and elements never seen before.

The kind of systems that seem to be especially adapted to on-the-job learning are dialogue systems, since they can
take advantage of their interactions with users and collect feedback in order to adapt and improve their components
over time.
Two categories of dialogue systems can be described: task-oriented dialogue systems and conversational dialogue systems.
A conversational dialogue system aims at finding the most accurate answer to a user's utterance in an open domain (chit chat).
A task-oriented dialogue system aims at fulfilling user's needs relative to a specific domain, like
finding information or elements given some criteria (i.e. restaurant hours, a recipe) or performing a specific task (i.e. booking a flight).
Such a system is usually composed of different components which can be grouped in the following categories : the language abilities, the strategy and the knowledge. 
Each of these components could be improved in a lifelong learning scenario.

Some dialogue systems performing on-the-job learning have been built and evaluated but no general methodology has been
defined yet (see Section \ref{sec:sota}).
Moreover, only few systems making their Natural Language Understanding (NLU) component learn on-the-job have been
built, although this step is the first of the dialogue system and is crucial for correctly achieving the task. 
Thus we describe in this paper the following contributions:

\begin{itemize}
    \item a first attempt at a general methodology to evaluate on-the-job learning dialogue systems.
    \item a task-oriented dialogue system which improves on-the-job its NLU component by collecting and inferring new training examples thanks to its interactions with users.
    \item an evaluation framework applying the described methodology on the previous NLU component with user simulation and Knowledge Base (KB) completion simulation to demonstrate the described evaluation methodology.
\end{itemize}{}

\section{Related Work}
\label{sec:sota}

\citet{bing_on_the_job_2020} described the three main steps of on-the-job learning under the machine learning paradigm.
These three steps can be extended to a system composed of different components (which can rely or not on machine/deep
learning algorithms) and thus be defined as following: 1) detect when a new piece of knowledge can be learned, 2) retrieve
and identify the new piece of knowledge and 3) adapt the component associated to the new piece of knowledge.
As illustrated in this section, the type of the new piece of knowledge and the methods that can be used to perform these three steps depend on the component to improve.

\citet{Mazumder2019BuildingAA} focused on improving the understanding component and built an application independent natural language interface that matches user's commands to
patterns (e.g. ``draw a X1 circle at X2'') associated to a specific action (Natural Language
to Natural Language).
Their approach relies basically on learning on-the-job new patterns.
To detect that a new piece of knowledge can be learned, their system explicitly asks if it correctly understood the
user, by only taking advantage of negative examples.
The user tells explicitly what was the correct requested action, so that the system can extract a pattern from the
initial user's query so that a new paraphrased command can be added to the system.
The learning process is thus continuous but under the closed world assumption.

To evaluate their different system's versions - including ones that did not learn on-the-job, they collected user's commands and computed the accuracy on the associated system's answers. However this does not allow the comparison over time of different system's states since the data used to make the system learn on-the-job (production) and evaluate it are the same.

A Knowledge Base (KB) can also be improved on-the-job as shown by \citet{mazumder-etal-2019-lifelong}. The authors developed an engine to help dialogue systems improve their factual KB thanks to their interactions with users.
When the user asks a question referring to unknown elements (relation or entity), the system asks the user for an
example of fact containing the unknown elements.
These new facts allow the system to infer other new facts which can help answer the initial question.
This way of improving the KB through this inference process is close to the definition of LL.
However, the authors decided to focus on the inference engine and directly worked with triple queries instead of natural language, which greatly simplify the detection step and the
retrieval of new elements to learn.
To simulate the production environment, they made use of a simulated user which can ask queries with unknown relations and entities.
Then they evaluated their system on the sames queries at the end of the production phase, meaning that the evaluation process is not continuous. Moreover they did not compare their system with an initial one that did not learn on-the-job.

On the other hand, \citet{hancock-etal-2019-learning} built a conversational dialogue system which collects new training examples
during conversation with its users and which periodically adapts its models thanks to these new training examples
(not a continuous on-the-job learning process).
They first perform the detection step by inferring the user satisfaction.
If the user seems satisfied, the system stores the dialogue as a new training example (imitation).
Otherwise it asks the user to give a more appropriate answer and the system's answer is replaced in the dialogue
by the answer provided by the user so that it can be used as a new training example.
To make their system learn on-the-job, they simulated a production environment on a crowdsourcing platform and
evaluated different variants of their system on a separated test dataset at the end on the learning process.
Thus they did not evaluate their models in a continuous manner.

Regarding the LL evaluation methodologies, \citet{Chen:2016:LML:3086758} described a commonly used evaluation methodology
for lifelong machine learning.
This methodology does not propose a continuous evaluation either and consists in a general evaluation.

Since there is no consensus about how to evaluate on-the-job learning dialogue systems, we state that a general
methodology needs to be defined. Thus we describe in the next section a first attempt at such an evaluation methodology, which takes the \emph{continuous} aspect into account and clarifies the difference between data used for the production and the evaluation phases.
Since no work has been done on improving on-the-job the slot-filling task independently, we describe and give the
evaluation results of a task-oriented dialogue system which improves on-the-job this task in the cooking domain in Section \ref{sec:eval_results}.

\section{General Methodologies to Evaluate On-the-job Learning Dialogue Systems}
\label{sec:eval_method}

We consider a dialogue system that can continuously and autonomously adapt one of its components on one specific task while interacting with users in production (on-the-job learning). The task itself would stay the same over time but the scope or even the domain could evolve.
We consider that the system has access to initial data, resulting in the initial state of the system along with an initial performance (e.g. trained model with the initial data). The system is then exposed to unknown situations and elements while being used in production (open environment) and should adapt to them over time, leading to different system states.

The main questions considering such a system are: (1) Does the system forget what it initially learned/knew while being used and while adapting itself - called catastrophic learning in continual learning \citep{Parisi2019ContinualLL}? (2) Does the system actually learn/accumulate the new elements that appear while being used, does it improve when adapting itself? (3) Does the system manage to infer additional knowledge from what it learned/accumulated while being used, in order to generalize?
In order to be able to truly compare all different system states with each other, the performance of each system state on the specified task has to be evaluated by always using the same test dataset. This dataset is actually composed of three different datasets: $test_{INITIAL}$: data that look like the initial data, in order to evaluate the feature (1); $test_{LEARN}$: data that contain elements that the system is supposed to learn/accumulate when being used, in order to evaluate the feature (2); $test_{UNKNOWN}$: data that contain elements that are not in the initial data and that are not going to appear when the system will be used, in order to evaluate the feature (3).

Additionally, the system adaptation has to be \emph{autonomous}, that means that the system has to adapt without the help of domain experts or developers. Knowing that, the only inputs accepted during the evaluation are interactions with end users, in natural language.
Moreover, in a real scenario it is not possible to define beforehand $test_{LEARN}$ and $test_{UNKNOWN}$. That is why the evaluation is based on user simulation - a common method to evaluate dialogue systems \citep{Deriu2020SurveyOE}) - in order to simulate the production phase. Thus, the data used for the simulation have to be similar to $test_{INITIAL}$ and $test_{LEARN}$ but be separate from $test_{UNKNOWN}$.

The system's \emph{continuous} learning also has to be evaluated.
In fact it could be problematic if the user had to correct the system for the same error again and again, waiting for the system to adapt.
Moreover, interactions with users and feedback collection should seem as natural as possible for the user. It is necessary if we want the adaptation process to be successful and keep the user willing to help the system to improve.
As a consequence, the system performance has to be evaluated after each interaction with a user.

To summarize, the evaluation methodology consists in simulating the user interactions and evaluating after each interaction the current system's performance on $test_{INITIAL}$, $test_{LEARN}$ and $test_{UNKNOWN}$.
We believe this methodology can be adapted to any on-the-job learning dialogue system in any domain.

\section{A Task-oriented Dialogue System Improving On-the-job its Understanding Component}
\label{sec:system}

\begin{figure*}
    \centering
    \includegraphics[scale=0.5]{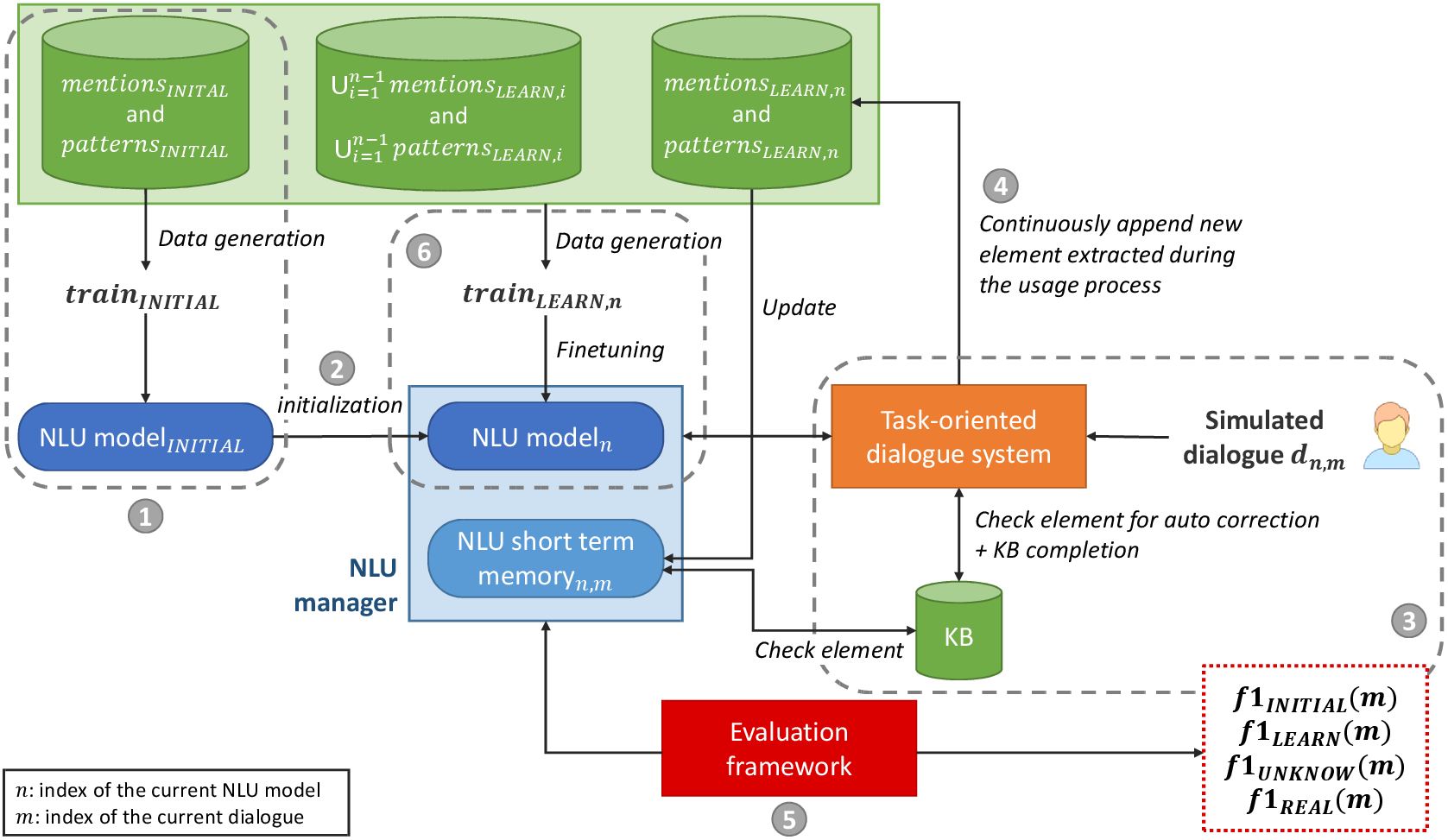}
    \caption{System operation and evaluation during production (simulated). 1) Training of the initial model on $train_{INITIAL}$, 2) Initialization of the NLU model, 3) Interactions with users, 4) Continuous extraction of new pieces of knowledge (mentions, patterns, training examples) and update of the STM, 5) evaluation on $test_{INITIAL}$, $test_{LEARN}$, $test_{UNKNOWN}$ and $test_{REAL}$ at the end of each dialogue and 6) If one of the conditions for model adaptation is validated, generation of $train_{LEARN, n}$ from past mentions/patterns and from new mentions/patterns and model fine-tuning.}
    \label{fig:schema}
\end{figure*}

The system that we built is a task-oriented dialogue system in the cooking domain which can continuously and autonomously improve when interacting with users (on-the-job learning). The operation of the system is described in Figure~\ref{fig:schema}.

\subsection{Description of the Task and the Model}

In this paper, we focus on the improvement of the Natural Language Understanding (NLU) component. NLU is usually composed of two steps: slot-filling and intent detection, but for first experimentation we decided to consider only slot-filling.
Slot-filling can be considered as a sequence labelling task which aims at retrieving from a user's utterance a set of concepts. A concept correspond to a tuple (type of concept, mention).
For the first experimentation we decided to work on the recipe domain and defined the following 8 types of concepts to detect: recipe type, ingredient, preparation technique, origin, origin adjective, meal, event, other category.
For instance, given the following user query \texttt{"I'd like to prepare chocolate cake for my son's birthday"}, the concepts to detect are \texttt{(ingredient, 'chocolate')}, \texttt{(recipe type, 'cake')} and \texttt{(event, 'birthday')}.
To take negative contexts into account we added the negative variant of each concept (e.g. for \texttt{"I don't like bananas"} the concept to detect is \texttt{(negative ingredient, 'bananas')}).

Slot-filling is often performed using deep neural networks. We decided to use a BiLSTM architecture which is known to achieve good performances on classical slot-filling tasks \citep{Bechet2018}.
Before the simulated production phase, we trained a double layered BiLSTM\footnote{Model hyperparameters: Adam optimizer with initial learning rate of 0.001, Word2Vec size 300, CrossEntropy loss.} with layers of size 128 on $train_{INITIAL, TRN}$ (see Section \ref{data_preparation}). The embedding layer is initialized with Word2Vec CBOW~\citep{mikolov2013distributed} vectors pre-trained on Wikipedia \citep{9053278}. Model selection was done by taking the highest F1-score on the development set ($train_{INITIAL, DEV}$, see Section \ref{data_preparation}).

\subsection{Collecting New Pieces of Knowledge}

To build this on-the-job learning dialogue system, the main idea is to take advantage of interactions with users to collect new training examples to adapt the NLU model.
The first step for collecting new pieces of knowledge is to \emph{detect} that one can be learned.
This step is performed by detecting a posteriori that an NLU error occurred  \citep{hancock-etal-2019-learning}: first the user addresses the system which answers him/her and tells him/her the concepts it detected (NLU output); if the user continues the conversation, the NLU output is supposed to be correct - principle of \emph{progessivity} \citep{Albert2018RepairTI}; if the system detects that the user is notifying that it misunderstood him/her, the NLU output is considered as incorrect \footnote{To perform this first step we used basic regular expressions (e.g. \texttt{r".*wrong.*"}, \texttt{r".*not what I.*"}).}.

The next step is the extraction and the identification of the new piece of knowledge, consisting here in labelling the initial user query.
If the NLU output is correct, the initial user query is labelled with the concepts detected and stored.
If the NLU output is incorrect, the system tries to correct itself. This correction is then proposed to the user: as previously, if the user keeps going, he/she implicitly confirms the correction, if not it means that the new piece of knowledge cannot be extracted.
To correct the initial NLU output we make the assumption that the user will rephrase his/her initial query after notifying the system that it misunderstood him/her (e.g. ``You misunderstood me, I asked for a cake recipe without eggs''). Note that in practice it is not always true \citep[p.29]{hough2014modelling}.
Under this assumption, the system compares the initial user query and the paraphrase by removing stop words and keeping only the common chunks (e.g. for the initial query \texttt{"Do you have cake recipes for people allergic to eggs?"} and the previous paraphrase, the common chunks are \texttt{cake} and \texttt{eggs}).
Then it checks in the KB associated to the dialogue system if an element associated to the chunks exists and tries to extract the type of concept associated. If it manages to extract the type, the initial user's query can be labelled. This step is crucial but depends highly on the quality of the KB and on its capability to be adapted to the concepts to detect.

Since learning on-the-job involves an open world environment, new elements that are not originally in the KB may appear over time. This can impact the performance of the NLU model since it will have to detect new mentions it never saw during the initial training. The KB should thus evolve over time to mitigate this issue.
We decided to incorporate this phenomenon in our experiments. To do it we intentionally removed some elements from the KB and simulate the completion of the KB when the system detects that the user refers to an element missing in the KB. In a real case, we can consider that this completion would be performed by extracting information about the missing element from the web for instance.
Moreover, since the labelled training examples collected during interactions with users would be too few to adapt the model, we decided to make the system inferring additional knowledge to augment the number of labelled training examples. To do it the system uses a basic method based on patterns. For instance, for the user's query \texttt{"Tonight I have a barbecue, can you suggest me something to prepare?"} with the concept \texttt{(event, 'barbecue')}, the pattern \texttt{"Tonight I have a \$event, can you suggest me something to prepare?"} can be extracted. The system can then generate multiple labelled training examples from this new pattern.
The new pieces of knowledge that can be collected consist thus in new mentions, new patterns or new training examples (if the pattern if not new).

\subsection{Adapting Natural Language Understanding}

Since the \emph{continuous} learning aspect is important and since a neural network needs time to be adapted, a short term memory which can directly use the knowledge learned from past dialogues has been added to the NLU component. To the best of our knowledge, it is the first time a short term and a long term memory (in our case the adapted model) have been jointly used for learning on-the-job.
The NLU component can be then adapted in two ways: a continuous adaptation with the Short Term Memory (STM) or an incremental adaptation with the adaptation of the NLU model.
In this experiment, only new mentions and their associated concept type are considered in the STM. In fact, we consider that a user will expect the system to directly remember new mentions introduced in past dialogues and that he/she will consider that the dialogue system is not learning on-the-job if it is not able to do it. This is less true for patterns.
Note that when the model is adapted, the STM is cleared.
After detecting the concepts in the user query thanks to the NLU model, the output is updated through the STM as described in Algorithm~\ref{alg:nlu_update_stm}. The strategy followed has been defined empirically through preliminary experiments. We observed in particular that taking into account the negative context really impacts the NLU update.

To adapt the NLU model, the current model is fine-tuned on the new labelled training examples extracted during interactions with users ($train_{LEARN, n}$ on Figure \ref{fig:schema}). The adaptation process is triggered if one of the conditions about the number of new mentions, new patterns or new examples is validated. We decided to use replay to prevent catastrophic forgetting \citep{Parisi2019ContinualLL} so that $train_{LEARN, n}$ is generated from both patterns and mentions from initial training and past adaptations and from the current new mentions and patterns. For each adaptation the system always generates 1000 training examples and adds the new examples collected during the interactions. For the fine-tuning, a new model $model_{n}$ is initialized with the weight of the previous model $model_{n-1}$ and trained on $train_{LEARN, n}$ \footnote{The parameters are the same than for the initial training except for the epoch size which is set to the length of $train_{LEARN, n}$. The same development dataset is also used.}.
Two kinds of replay are tested: 1) Replay on Patterns and Mentions (RPM) and 2) Replay on Mentions only (RM).
 
\begin{algorithm}
\caption{Natural Language Understanding update strategy through Short Term Memory}\label{alg:nlu_update_stm}
\hspace*{\algorithmicindent} \textbf{Input:} user's utterance u \\
\hspace*{\algorithmicindent} \textbf{Output:} concepts detected in u
\footnotesize
\begin{algorithmic}[1]
\Procedure{get\_concepts}{$u$}
\State $concepts_{model} = get\_concepts\_from\_model(u)$
\State $concepts_{stm} = get\_concepts\_from\_stm(u)$
\State initialize $concepts$ as an empty list of concepts
\If{$concept_{stm}$}
    \For{$c_{model}$ in $concepts_{model}$}
        \For{$c_{stm}$ in $concepts_{stm}$}
            \If{$c_{model}.mention == c_{stm}.mention$ and $c_{model}.type == c_{stm}.type$}
                \State add $c_{model}$ to $concepts$
            \ElsIf{$c_{model}.mention == c_{stm}.mention$ and is\_negative($c_{model}.type$)}
                \State add (negative($c_{stm}.type$), $c_{stm}.mention$) to $concepts$
            \ElsIf{$c_{model}.mention$ in $c_{stm}.mention$}
                \State add $c_{stm}$ to $concepts$
            \ElsIf{$c_{stm}.mention$ in $c_{model}.mention$}
                \If{$c_{model}.mention$ in KB}
                    \State add $c_{model}$ to $concepts$
                \Else
                    \State add $c_{stm}$ to $concepts$
                \EndIf
            \EndIf
        \EndFor
        \If{nothing added to concepts}
            \State add $c_{model}$ to $concepts$
        \EndIf
    \EndFor
\Else
    \State $concepts = concepts_{model}$ 
\EndIf
\State \textbf{return} $concepts$
\EndProcedure
\end{algorithmic}
\end{algorithm}

\section{Evaluation Framework and Results}
\label{sec:eval_results}

\subsection{User Simulation}

A stated in Section \ref{sec:eval_method}, a user simulation is used to simulate the production phase in order to evaluate the system.
The simulated user consists here in a ruled-based program. Only the first dialogue utterance is generated (the set of first utterances corresponds to the dataset $simulation$).
The simulated user follows the following scenario: 1) the user asks for a recipe by providing some criteria (i.e. the type of recipe, the ingredient, etc) 2) the system provides an answer and gives the criteria it detected (the concepts) 3) if the criteria detected by the system don't match the ones of the user, he/she tells the system \texttt{"You misunderstood me. I want a recipe with \$concepts."} with the concepts in natural language in place of \$concepts, otherwise he/she behaves normally 4) if it was wrong the system provides a corrected answer and gives the criteria it corrected 5) if the corrected concepts match the user's criteria he/she behaves normally, otherwise he/she tells goodbye \footnote{The decision diagram describing the user's and the system's behaviour can be found in the appendix.}.
Note here that with this simple user simulation, the first step of on-the-job learning - which consists here in detecting that the system misunderstood the user - becomes trivial.

\subsection{Data Preparation and Evaluation}
\label{data_preparation}

To train, simulate and evaluate our on-the-job learning dialogue system the following datasets are needed: $train_{INITIAL, TRN}$ (20k), $train_{INITIAL, DEV}$ (4k), $simulation$ (20k), $test_{INITIAL}$ (1k), $test_{LEARN}$ (1k) and $test_{UNKNOWN}$ (1k).
Because we initially had no data for the concept detection task, we decided to generate the data using patterns and mentions. The patterns were written manually by two people from the research domain and the mentions were scraped from the internet. Moreover, generating data from patterns and mentions allowed us to have more control on the new pieces of knowledge to learn during the simulated production phase.
The sets of patterns and mentions are randomly split into disjoint sets, namely $INITIAL$, $LEARN$ and $UNKNOWN$ to then generate the respective datasets.
For $simulation$ we made the assumption that the probability to have a new pattern in a user's query is equal to 0.7 and that the probability to have a new mention is equal to 0.3 (a user can convey the same intention using plenty of different sentences construction but have less variety in the mentions he/she uses). The $simulation$ dataset is built from $INITIAL$ and $LEARN$ with a mention and patterns distribution following this assumption \footnote{For the first experiment, we removed from the KB around 40\% of the less frequent ingredients so that the unknown mentions in $LEARN$ only consists in unknown ingredient mentions.}.
For the development dataset, we made the same assumption and also split the set $INITIAL$ into to disjoint sets so that $train_{INITIAL, DEV}$ consists in some mentions and patterns known in $train_{INITIAL, TRN}$ and some others that are unknown in order try to select the model which is the best at generalizing in an open world environment.
Since all the datasets have been generated, we added a real test dataset $test_{REAL}$ (744 annotated queries), composed of questions relative to the cooking domain \footnote{collected as part of the ERA-Net CHIST-ERA LIHLITH project.}.

To evaluate the performance of the NLU component, the F1-score is computed with the conlleval.pl script \citep{tjong-kim-sang-buchholz-2000-introduction} at the end of each simulated dialogue after the component's adaptation. 
To assess if the initial random splits for the data preparation impact the results, we prepared 4 different set of datasets with different random seed values. For each set of datasets an initial NLU model is trained and 2 simulations with the two kinds of replay are conducted.

\subsection{Results and Discussion}
\label{results}

\begin{figure*}
    \centering
    \includegraphics[scale=0.4]{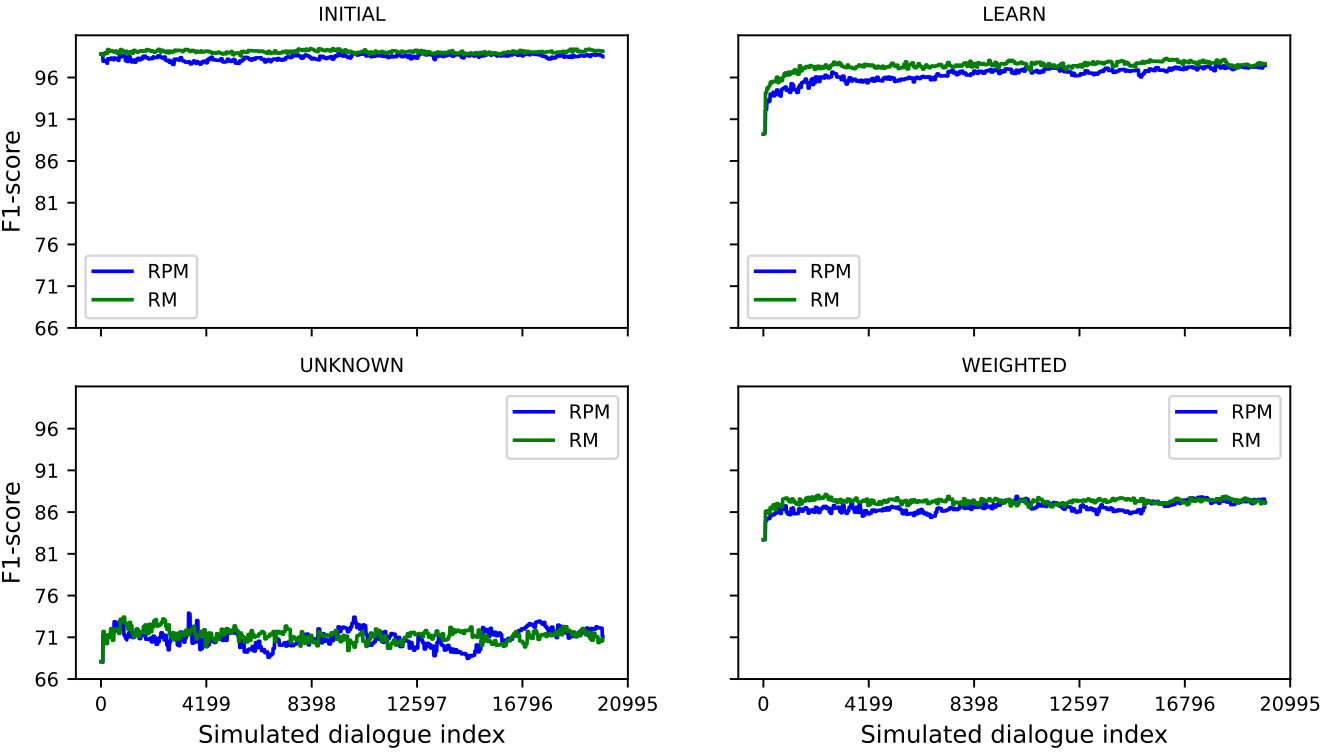}
    \caption{F1-score evolution during simulation on $test_{INITIAL}$, $test_{LEARN}$, $test_{UNKNOWN}$ and $test_{WEIGHTED}$ with data prepared with random seed 1 ($F1_{WEIGHTED} = F1_{INITIAL}*0.2 + F1_{LEARN}*0.4 + F1_{UNKNOWN}*0.4$).}
    \label{fig:f1_evolution}
\end{figure*}

Figure \ref{fig:f1_evolution} shows the evolution of the F1-score during simulation \footnote{We do not show the F1-score evolution for the experiments with data prepared with other random seed values since the behaviour is the same.}.
We did not observe significant performance differences when the system adapts its component using Replay on Patterns and Mentions (RPM) or using Replay with Mentions only (RM).
On $test_{LEARN}$ both F1-scores increase significantly at the beginning of the simulation suggesting that learning new mentions is in our case more profitable for the system when adapting its component (since $simulation$ is generated randomly there are more new mentions to learn at the beginning of the simulation).
However we can see that RM is performing better than RPM (about 2 more points) and that RPM increases progressively until achieving similar F1-score than RM at the end of the simulation. This again suggests that learning new mentions is more profitable since the new mentions will be two times less represented in $train_{LEARN, n}$ for RPM compared to RM.
Moreover, the fact that they both seem to reach a plateau shows the limitations of an experimental protocol relying on generated data.

Since the count of fine tunings during the simulation is high (322/327 for RPM/RM), the continuous evolution is not visible on this figure.
We computed the F1-score difference between each model fine-tuning to better analyse the impact of the STM. We observed that during the simulation the STM improves the F1-score of maximum 0.43 points (between simulated dialogues 318 and 390 with 16 new mentions) and reduces the F1-score of a maximum of 0.04 points on $test_{LEARN}$ with RPM (mean: 0.01, median: 0).

The Table \ref{tab:results} shows the F1-scores of different models on the test datasets.
The results for $model_{LEARN, RPM, n=N}$ and $model_{LEARN, RM, n=N}$ compared to $model_{INITIAL}$ show that the NLU component is improving with the simulated interactions thanks to the collection and adaptation methods described in Section \ref{sec:system}.
Comparison with $model_{LEARN, STM}$ shows that the STM alone is not sufficient.
When comparing with $model_{SIMU}$ we observe similar scores except on $test_{LEARN}$ and $test_{REAL}$. We suppose that it comes from the fact that all the new pieces of knowledge available in the simulated dialogues could not be extracted by the system. From the system's log we observe in particular that only 0.14 percent of the user's initial queries could be correctly annotated after the user rephrases his/her query because the system misunderstood him/her. This shows that the KB was not completely adapted for the concepts that have to be detected.

\begin{table*}
    \centering
    \scriptsize
    \begin{tabular}{|c||c|c|c|c|c|}
        \hline
          & $test_{INITIAL}$ & $test_{LEARN}$ & $test_{UNKNOWN}$ & $test_{WEIGHTED}$ & $test_{REAL}$\\
        \hline\hline
        $model_{INITIAL}$ & 98.99 & 89.33 & 68.26 & 82.83 & 36.18 \\
        \hline
        $model_{LEARN, STM}$ & 98.50 (-0.49) & 91.61 (+2.28) & 67.52 (-0.74) & 83.36 (+0.53) & -\\
        \hline
        $model_{LEARN, RPM, n=N}$ & 98.56 (-0.43) & 97.48 (+8.15) & 71.11 (+2.85) & 87.15 (+4.32) & 37.16 (+0.98)\\
        \hline
        $model_{LEARN, RM, n=N}$ & 99.14 (-0.15) & 97.63 (+8.30) & 70.79 (+2.53) & 87.20 (+4.37) & 37.70 (+1.52)\\
        \hline
        $model_{SIMU}$ & 99.94 (+0.95) & 99.60 (+10.27) & 71.49 (+3.23) & 88.42 (+5.59) & 41.24 (+5.06)\\
        \hline
    \end{tabular}
    \caption{F1-scores on the test datasets with data prepared with random seed 1. N is the index of the last fine-tuning at the end of the simulation (here N=322/327 for RPM/RM). $model_{LEARN, STM}$ corresponds to $model_{INITIAL}$ adapted only with the STM (no fine-tuning during simulation). $model_{SIMU}$ corresponds to $model_{INITIAL}$ fine-tuned on the $simulation$ dataset.}
    \label{tab:results}
\end{table*}

Considering the fact that simple methods have been implemented to perform the three steps of on-the-job learning - as described in Section \ref{sec:system} - we obtain promising results suggesting that using more sophisticated methods could lead to generalisation capabilities demonstrating the possibility of a truly LL task-oriented dialogue system.

\section{Conclusion and Future Work}

In this paper we described a first attempt at a general methodology to evaluate in a continuous manner the capability of a dialogue system to learn on-the-job. To the best of our knowledge it is the first general methodology defined of this kind. This methodology consists basically in simulating the user interactions and evaluating after each interaction the current system's performance on the following test datasets: $test_{INITIAL}$ (data similar to initial ones), $test_{LEARN}$ (data that contain elements that the system is supposed to learn during simulation) and $test_{UNKNOWN}$ (data that contain elements that are not in the initial data and that are not going to appear during simulation). We also built a task-oriented dialogue system which can continuously and autonomously improve its understanding component thanks to its interactions with users and evaluated it thanks to user simulation according to the described evaluation methodology.
Our methodology enabled us to compare \emph{over time} the different adaptation methods and identify how they were different, whereas the evaluation protocols used in other works could not have allowed this analysis.

For future work we plan to adapt this experiment to other domains and to add to the evaluation methodology a way to evaluate the system's robustness to noise (e.g. user providing a wrong correction).

\section*{Acknowledgments}

This work has been supported by a CIFRE convention, funded by ANRT (France, convention 2019/0628), and by ERA-Net CHIST-ERA LIHLITH Project, funded by ANR (France, project ANR-17-CHR2-0001-03).


\bibliographystyle{apalike}
\bibliography{main}

\clearpage

\appendix

\section{Simulated User and System Behaviour}

\begin{figure}[h!]
    \centering
    \includegraphics[scale=0.67]{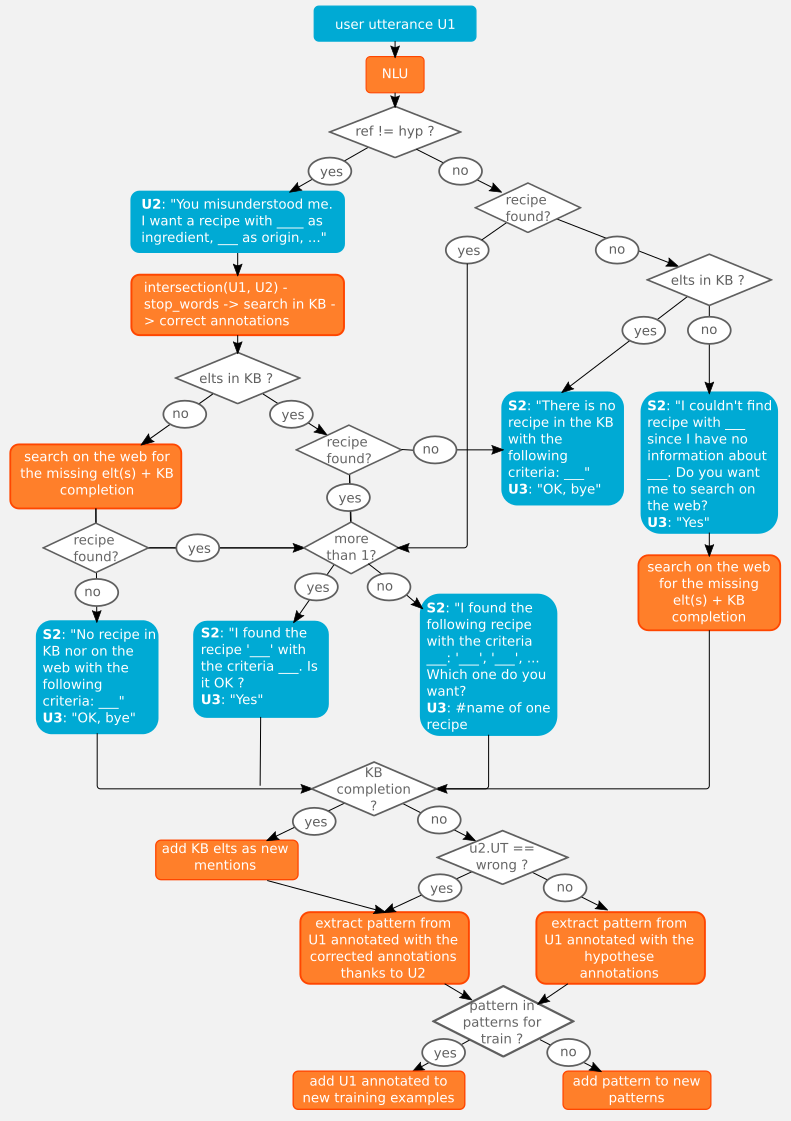}
    \caption{Decision diagram describing the user's and the system's behaviour. The text in blue boxes denotes user's $Ui$ and system's $Si$ utterances with $i$ the index of the turn. Text in orange boxes denotes system's internal operations.}
    \label{fig:my_label}
\end{figure}

\end{document}